\documentclass[conference]{IEEEtran}
\usepackage{amsmath,amsfonts}
\usepackage{array}
\usepackage{textcomp}
\usepackage{url}
\usepackage{graphicx}
\usepackage{cite}
\usepackage{float}

\usepackage{booktabs}
\usepackage{makecell}
\usepackage{enumerate}

\usepackage{hyperref}
\hypersetup{hypertex=true,
colorlinks=true,
linkcolor=blue,
anchorcolor=blue,
citecolor=blue}

\begin{document}

\title{On-Meter Graph Machine Learning: A Case Study of PV Power Forecasting for Grid Edge Intelligence} 

\author{\IEEEauthorblockN{Jian Huang, Zixiang Ming, Yongli Zhu*, Linna Xu}
\IEEEauthorblockA{Sun Yat-sen University, Guangzhou, China (yzhu16@alum.utk.edu)}}



\maketitle

\begin{abstract}

This paper presents a detailed study of how graph neural networks can be used on edge intelligent meters in a microgrid to forecast photovoltaic power generation. The problem background and the adopted technologies are introduced, including ONNX and ONNX Runtime. The hardware and software specifications of the smart meter are also briefly described. Then, the paper focuses on the training and deployment of two graph machine learning models, GCN and GraphSAGE, with particular emphasis on developing and deploying a customized ONNX operator for GCN. Finally, a case study is conducted using real datasets from a village microgrid. The performance of the two models is compared on both the PC and the smart meter, exhibiting successful deployments and executions on the smart meter. \\

\end{abstract}


\section{Introduction}

\IEEEPARstart{I}{n} recent years, the widespread deployment of smart meters and concentrators has fundamentally transformed power grid operations. These edge-side assets enable ``edge autonomy" for distribution networks, allowing local devices to execute complex algorithms and reduce reliance on centralized cloud computing. This shift is the essence of the so-called ``grid edge intelligence".

The core of this trend is ``on-device inference", which is happening in many areas such as  IoT (Internet of Things), automotive, and energy sectors. For example, various predictive models, such as support vector machines, gradient-boosting trees, and linear regression, have been integrated into smart meters for a voltage regulation task \cite{ref1}. Related works have also explored running object-detection frameworks on low-power-consumption hardware \cite{ref2}. In \cite{ref3}, the author deploys machine learning models via MATLAB Coder on a smart meter to control the PV (photovoltaic) inverters' power output.

In many power grid applications, the data is inherently graph-structured, with stations as nodes and power lines as edges. Graph Neural Networks (GNNs) have become a state-of-the-art model for processing such data structures. In \cite{ref4}, the author proposed a GNN model named DEST-GNN to exploit the inherent spatio-temporal correlation with other PV stations to predict the power of each PV station simultaneously. In \cite{ref5}, the author uses Graph Neural Networks and Multi-Agent Reinforcement Learning for voltage regulation in RES-integrated grids.

In the field of deep learning, there are several mainstream training frameworks (such as PyTorch and TensorFlow), but their different model representation formats hinder cross-platform deployment and model migration. In response, the Open Neural Network Exchange (ONNX) format was proposed \cite{onnx_website}. It provides a unified, open standard for model representation to promote interoperability among frameworks. 

However, deploying GNN-related models on the resource-constrained meter encounters issues: certain GNN operators are missing from the ONNX built-in library, e.g., GCN (Graph Convolutional Network) layers \cite{GCN}. Furthermore, for certain built-in GNN layers, such as GraphSAGE\cite{GraphSAGE}, the ONNX computation graph generated by its official implementation differs from the original layer's behavior. These issues hinder on-meter model deployment. To address these issues, this paper modifies the specific models and extends the original operator set with custom implementations, ultimately successfully deploying GCN and GraphSAGE models on the meter for photovoltaic power forecasting. The main contributions are summarized as follows:
\begin{itemize}
    \item Develop a customized GCN operator for ONNX.
    \item Deploying a GCN-based PV forecasting model on the smart meter using our customized GCN operators.
    \item Deploying a GraphSAGE-based PV forecasting model on the smart meter, and comparing it with the GCN model.
\end{itemize}

The structure of this paper is as follows. Section \ref{sec2} describes the basic principles of GCN and GraphSAGE, the specification of the prediction task, the hardware specification of the meter, and key deployment tools such as ONNX and ONNX Runtime. Section \ref{sec3} elaborates the specific deployment process for the GCN and GraphSAGE models. Section \ref{sec4} presents and analyzes the performance of these two models on the meter.

\section{Problem Description and Background Theory}\label{sec2}

\subsection{Forecasting PV Power in a ``Cloud-less" Village Microgrid}

In remote microgrid scenarios, on-meter PV forecasting is crucial for local energy management, particularly when poor communication infrastructure deteriorates cloud connectivity. Due to unstable communication with the control centers and the absence of external grid support, forecasting must be performed locally without real-time meteorological data (e.g., solar irradiance). Thus, the forecasting model may only rely on historical power measurements as its inputs. This process is formulated in Eq. (\ref{eq5}), where $n$ is the number of stations, \(k\) denotes the input feature length per station, \(h\) is the prediction horizon, and $f$ stands for the GNN-based mapping function.

\begin{equation}
\label{eq5}
\begin{bmatrix} P_{1,t+h} \\ P_{2,t+h} \\ \vdots \\ P_{n,t+h} \end{bmatrix}=f(\begin{bmatrix}
P_{1,t} & P_{1,t-1} & \cdots & P_{1,t-k+1} \\
P_{2,t} & P_{2,t-1} & \cdots & P_{2,t-k+1} \\
\vdots & \vdots & \ddots & \vdots \\
P_{n,t} & P_{n,t-1} & \cdots & P_{n,t-k+1}
\end{bmatrix})
\end{equation}

\subsection{A brief introduction to GCN}

The essence of graph convolutional networks is to extract spatial features from the graph structure. Depending on the extraction method, they can be categorized into spatial-domain graph networks (e.g., GraphSAGE, GAT) and spectral-domain graph networks (e.g., Spectral CNN, GCN). This section will provide a brief introduction to GCN.

GCN is a deep learning model specifically designed to process graph-structure data. Unlike traditional convolutional neural networks that extract local features from regularly structured data such as images, GCNs define a convolutional operation on graphs. This operation effectively aggregates feature information from each node and its neighbors, enabling the model to learn the graph's topological structure and node attributes efficiently.

The core idea of a GCN is to update nodes' feature representations via a layer-wise propagation rule. For a given graph $G=(V, E)$, where $V$ is the set of nodes and $E$ is the set of edges, we can describe this graph using a node feature matrix $X \in R^{N \times F}$ and an adjacency matrix $A \in R^{N \times N}$. $N$ is the number of nodes, and $F$ is the feature dimension of the initial input for each node.

A multi-layer GCN model calculates the feature representation of nodes at the $l+1$-th layer, denoted as $H^{l+1}$, through the following layer-wise propagation rule:
\begin{equation}
H^{(l+1)} = \sigma \left( \tilde{D}^{-\frac{1}{2}} \tilde{A} \tilde{D}^{-\frac{1}{2}} H^{(l)} W^{(l)} \right)    
\end{equation}
where:
\begin{itemize}
    \item $H^l \in R^{N \times D_l}$ is the node feature matrix of the $l$-th layer, and $D_l$ is the feature dimension of this layer.
    \item $W^l \in R^{D_l \times D_{l+1}}$ is the trainable weight matrix of the $l$-th layer (the parameters that the model tries to learn).
    \item $\sigma$ is a non-linear activation function.
    \item $\tilde{A}$ is the adjacency matrix with added self-loops, defined as $\tilde{A} = A+I_N$, where $I_N$ is the identity matrix.
    \item $\tilde{D}$ is the diagonal matrix of $\tilde{A}$, where the elements on its diagonal are $D_{ii}=\sum_j\tilde{A_{ij}}$.
\end{itemize}

From a spectral analysis perspective, this propagation formula is a first-order approximation of spectral graph convolutions \cite{GCN}. By using a first-order approximation of Chebyshev polynomials, GCN simplifies complex spectral-domain operations into efficient spatial-domain neighbor aggregation, thereby improving the models' computational efficiency and scalability. 
A $k$-layer GCN model allows the final representation of each node to aggregate feature information from its $k$-hop neighbors, thereby learning rich structural representations that can be applied to downstream tasks such as node classification and graph classification.

\subsection{A Brief Introduction to GraphSAGE}
Like GCN, GraphSAGE is a graph neural network model whose core is to learn a function that generates a vector representation for a node by aggregating information from its neighbors \cite{GraphSAGE}. The entire algorithm can be broken down into two steps: neighbor aggregation and self-node update. If we want to generate an embedding for a node $v$ in the graph, the process iterates $K$ rounds, with each round gathering information from further neighbors. The iterative process for the $k$-th round is shown below.

Let $h_{N(v)}^k$ be the representation of the aggregated neighbor information for node $v$ at the $k$-th round, then we have:

\begin{equation}
    h_{N(v)}^k = AGGREGATE_k(h_u^{k-1}, \forall u \in N(v))
\end{equation}

In this paper, we use the following mean aggregation operation:
\begin{equation}
    h_{N(v)}^k = \frac{1}{|N(v)|}\sum_{u \in N(v)}h_u^{k-1}
\end{equation}
where:
\begin{itemize}
    \item $N(v)$ represents the set of neighboring nodes of node $v$.
    \item $|N(v)|$ represents the number of neighbors.
    \item $h_u^{k-1}$ represents the vector representation of a neighboring node $u$ at
    the $k-1$ round. In the first round (\textit{k}=1), $h_u^0$ is the node's initial input feature, $x_u$.
\end{itemize}

After obtaining the aggregated neighbor information, the model concatenates it with node \textit{v}'s own representation from the previous round, $h_v^{k-1}$, and then passes it through a fully connected layer and a non-linear activation function $\sigma$. The representation of node \textit{v} at the $k$-th round, $h_v^k$, is:
\begin{equation}
    h_v^k=\sigma(W^k \cdot \text{CONCAT}(h_v^{k-1},h_{N(v)}^k))
\end{equation}
where:
\begin{itemize}
    \item $W^k$ is the weight matrix to be learned in the $k$-th round.
    \item CONCAT($\cdot, \cdot$) denotes the concatenation of two vectors.
\end{itemize}

This process is iterated from $k$=1 to $K$. After completing $K$ rounds, the resulting $h_v^K$ is the final embedding representation for node $v$. Through this method, each node can integrate the structural and feature information from its $K$-hop neighbors.

\subsection{ONNX and ONNX Runtime}
ONNX serves as an intermediate representation, abstracting neural networks into a directed acyclic graph (called "computation graph") composed of nodes and tensors, where each node represents a specific operation (e.g., matrix multiplication, convolution). Through this standardized format, models from various frameworks can be converted to ONNX format.

\begin{figure}
    \centering
    \includegraphics[width=1\linewidth]{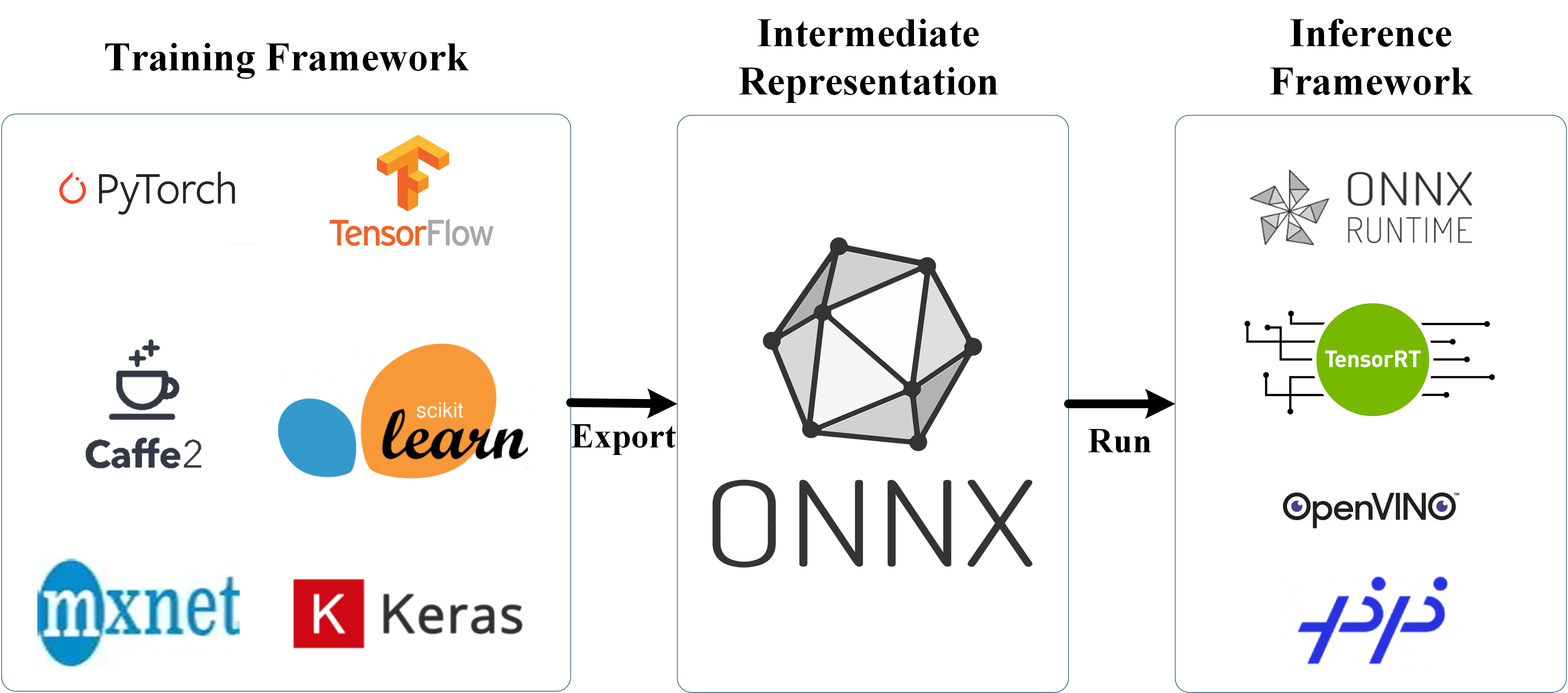}
    \caption{Illustration of ONNX and ONNX Runtime}
    \label{fig:onnx_func}
\end{figure}

ONNX Runtime is the official deployment and inference engine for executing models in the ONNX format. While ONNX defines the model's computation graph, ONNX Runtime loads exported ONNX models and efficiently executes inference across various hardware and operating systems \cite{onnx_runtime_website}. It provides a unified API (Application Programming Interface) that covers mainstream platforms and programming languages, enabling rapid deployment of models to specific hardware, as illustrated in Fig. \ref{fig:onnx_func}.

\subsection{Customized Operator}

Although the built-in operators provided by ONNX can handle most neural network models, some models, like GCN, contain operations that lack corresponding operator mappings. This prevents these models from being exported to the ONNX format. To solve this problem, ONNX provides a mechanism for creating customized operators, allowing developers to define new operators.

\begin{figure}
    \centering
    \includegraphics[width=1\linewidth]{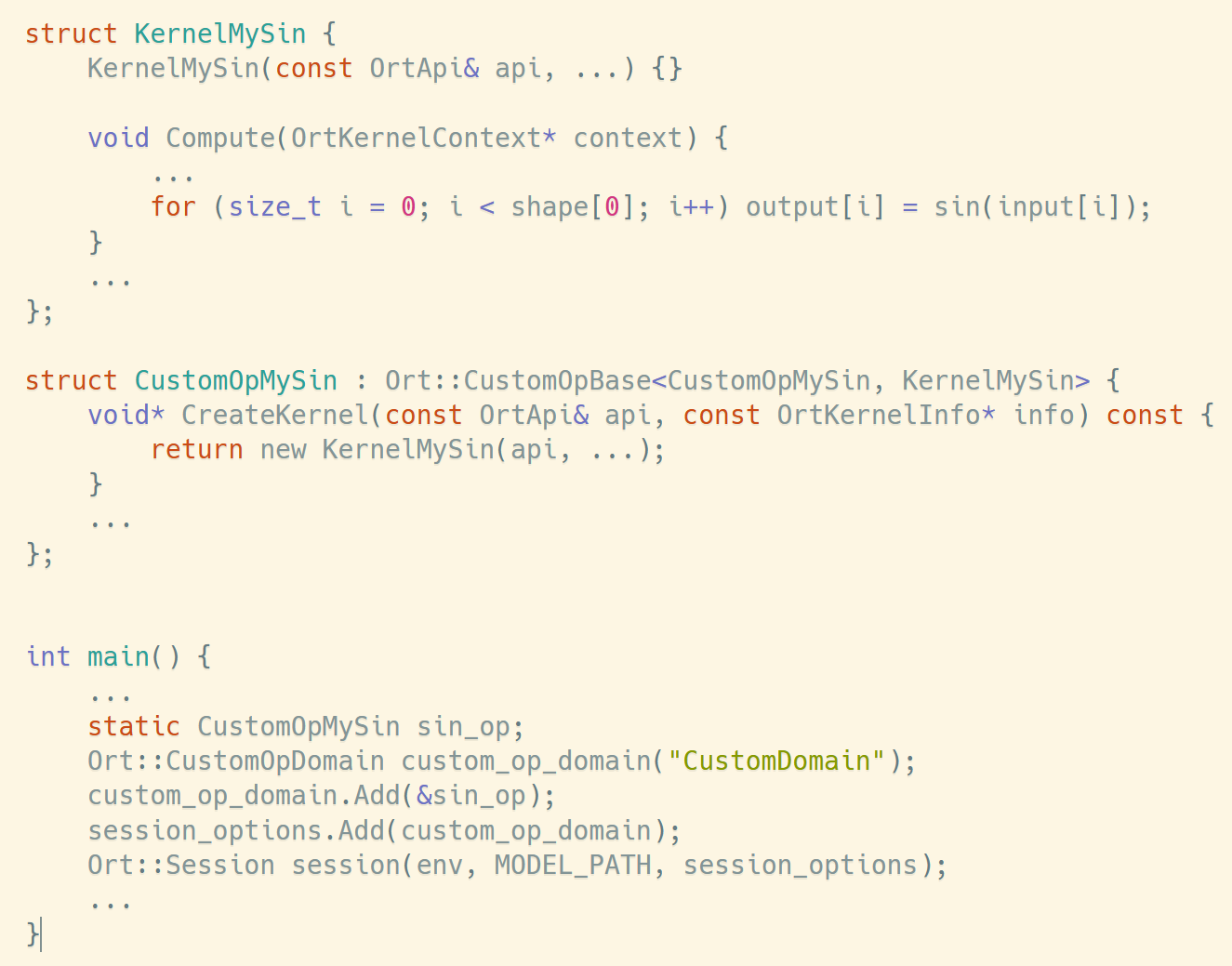}
    \caption{Backend kernel implementation}
    \label{fig:code3}
\end{figure}

Creating a customized operator involves the following three steps. Here we take the simple example of \textbf{\textit{sin}} function as an example of implementing a customized operator:

\begin{enumerate}
  \item \textbf{Define the operator schema}: The operator's inputs, outputs, and its forward and symbolic functions are defined in the training framework \cite{Torch_export}, as shown in Fig. \ref{fig:code1}.

  \begin{figure}[H]
      \centering
      \includegraphics[width=1\linewidth]{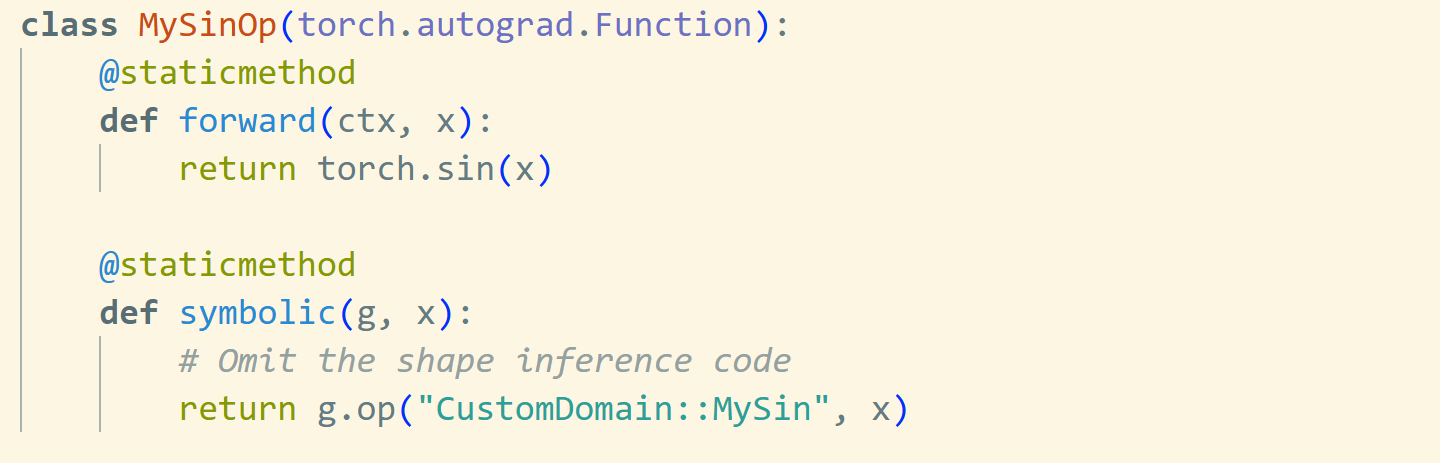}
      \caption{Define the operator schema}
      \label{fig:code1}
  \end{figure}
  
  \item \textbf{Model export integration}: During model export, the corresponding layer in the model is mapped to the customized operator defined above, as shown in Fig. \ref{fig:code2}.

\begin{figure}[H]
    \centering
    \includegraphics[width=1\linewidth]{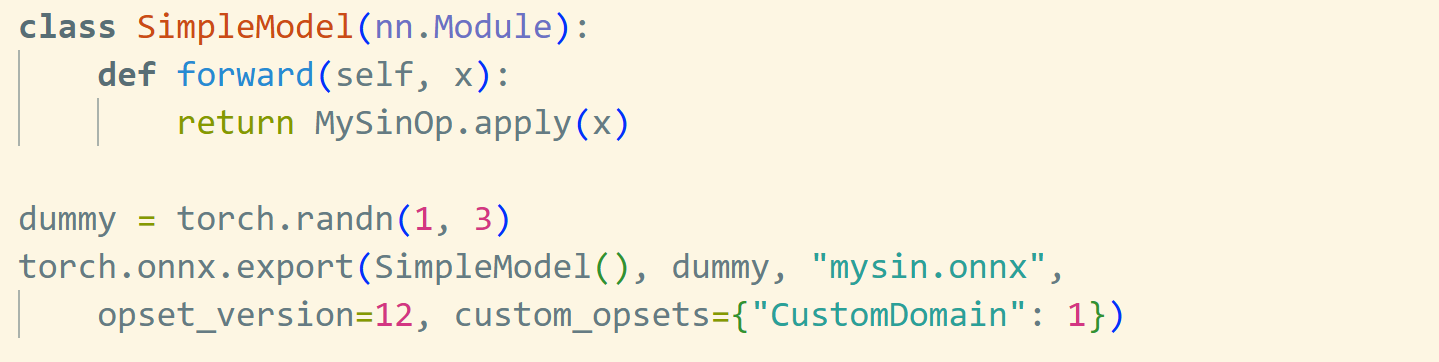}
    \caption{Model export integration}
    \label{fig:code2}
\end{figure}
  
  \item \textbf{Backend kernel implementation}: Implement the specific computational logic for the customized operator within the inference engine \cite{onnx_runtime_custom} (as shown in Fig. \ref{fig:code3}).

\end{enumerate}

This approach allows for the flexible development of more complex models/operators beyond ONNX's built-in library.

\subsection{Hardware and Software Specifications of the Smart Meter}

\begin{figure}
    \centering
    \includegraphics[width=1\linewidth]{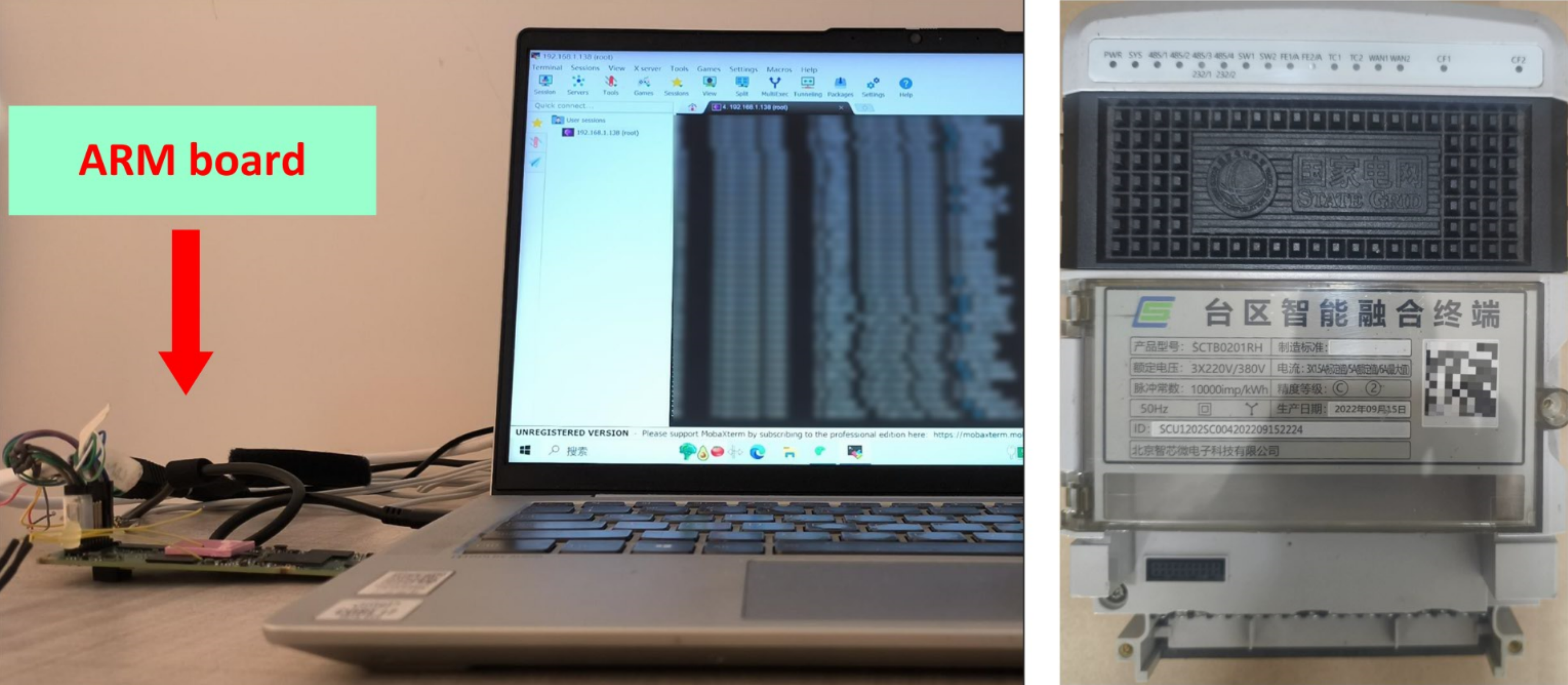}
    \caption{The real smart meter used in this paper}
    \label{fig:meter}
\end{figure}

The hardware platform for GNN deployment, depicted in Fig. \ref{fig:meter}, is an ARM-based, real smart meter. It is equipped with a quad-core Cortex-A53 CPU, 984 MB of RAM, and a 977 MB flash disk, of which 478 MB remains available for user applications after accounting for system overhead. The operating system (OS) is 64-bit ARM-Linux 4.9. It is worth mentioning that the adopted meter is a mass-production type that is \textit{not specially} designed for on-meter training.

\section{Deploying Graph Neural Network on the Smart Meter} \label{sec3}
In the previous sections, we have briefly introduced the theoretical foundations of GCN and GraphSAGE, as well as the hardware specifications of the target meter terminal. In this section, we will explain in detail the training and deployment process of these two models on the meter.

\subsection{Training and Deployment of GCN}\label{AA}

\begin{figure}
    \centering
    \includegraphics[width=0.95\linewidth]{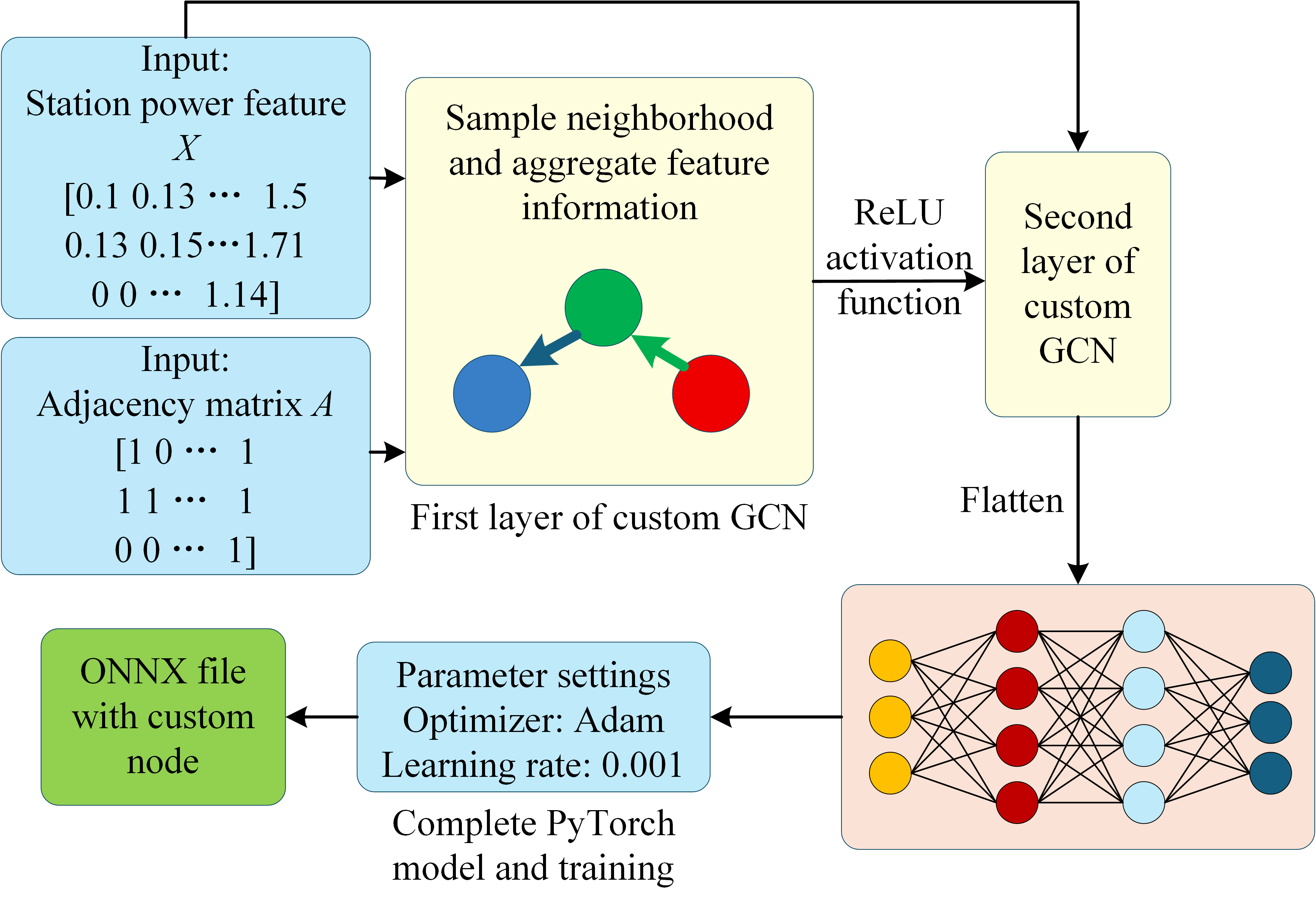}
    \caption{The architecture of the adopted GCN model}
    \label{fig:model_architecture}
\end{figure}

The neural network architecture used for photovoltaic forecasting is shown in Fig. \ref{fig:model_architecture}. The model is primarily composed of two GCN layers and one MLP layer. The model takes the historical power data of each station and an adjacency matrix as input. The GCN layers extract spatial dependencies among stations, while the MLP layer maps these high-dimensional features to the final prediction task.

After implementing this model, we discovered that the GCN layer in PyG (PyTorch Geometric) could not be converted directly to ONNX format. Therefore, we designed a customized GCN operator in PyTorch. It defines the computation logic via a forward function and provides the mapping rules for the ONNX format via a symbolic function. 

After replacing the original GCN layer with this customized operator and completing model training, we successfully export the model. When the obtained ONNX file is visualized, its computation graph is as shown in Fig. \ref{fig:gcn_onnx}. To resolve the inherent interpretability issues of raw ONNX graphs, we semantically group the operators into key architectural blocks. As shown in the figure, our customized GCN operator has successfully become an independent node (the \textbf{MyGcnOp} node) within the 2-Layer GCN Block, seamlessly connecting to the subsequent reshaping block and MLP block.

Although the model is successfully converted, the built-in library of ONNX Runtime does not contain the definition of our customized GCN operator, which prevents our ONNX file from being loaded and executed. Therefore, we need to provide a specific implementation for this customized operator within the ONNX Runtime framework.

\begin{figure}
    \centering
    \includegraphics[width=1\linewidth]{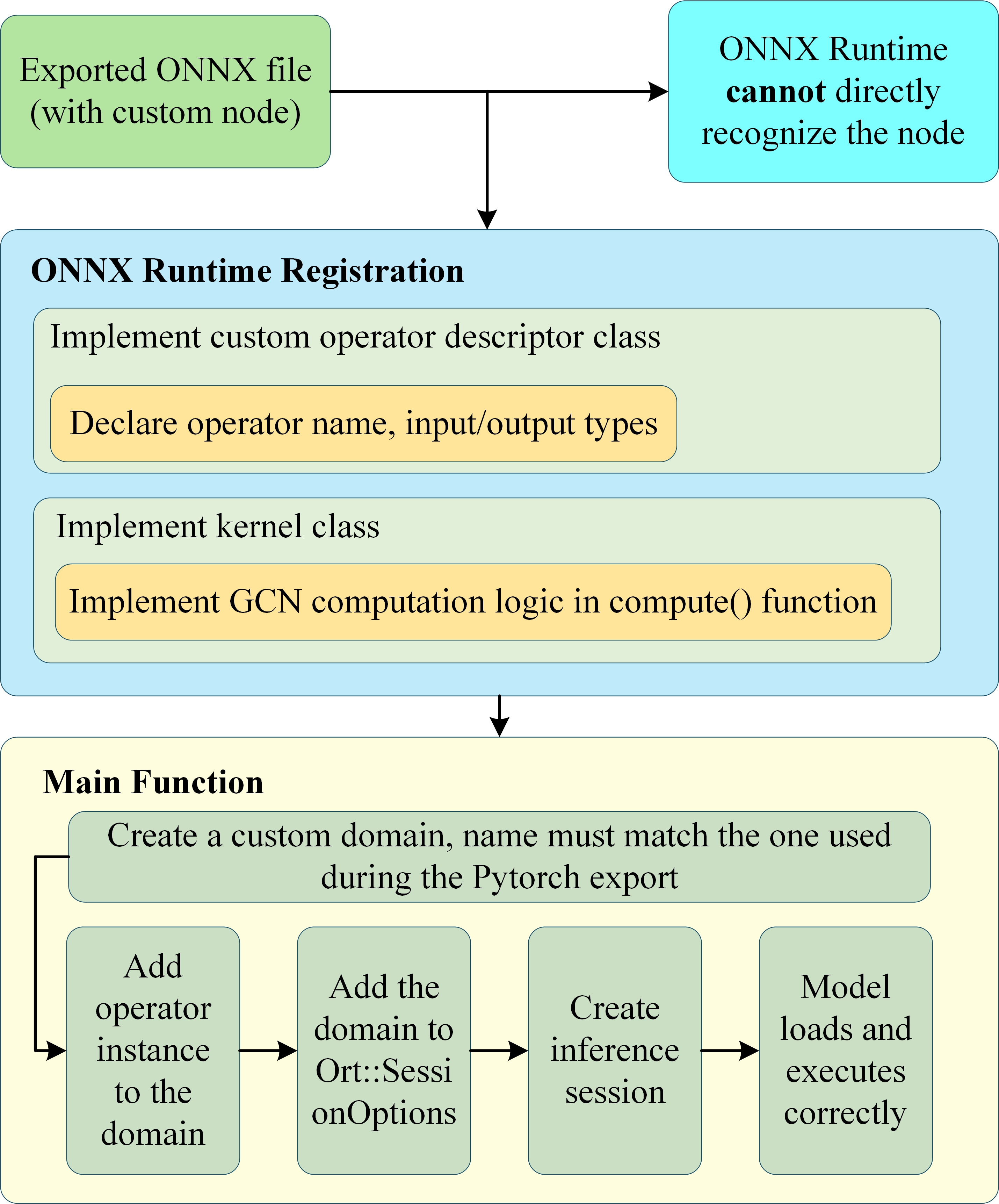}
    \caption{The process for loading models with customized GCN operators in ONNX Runtime}
    \label{fig:gcn_operator_ort}
\end{figure}

Fig. \ref{fig:gcn_operator_ort} details the entire process of implementing and registering this customized operator in ONNX Runtime. After completing these steps, the exported model can be correctly loaded and executed by the inference engine, thus enabling the deployment of a GNN model with our customized GCN operators on the smart meter.

\begin{figure}
    \centering
    \includegraphics[width=1\linewidth]{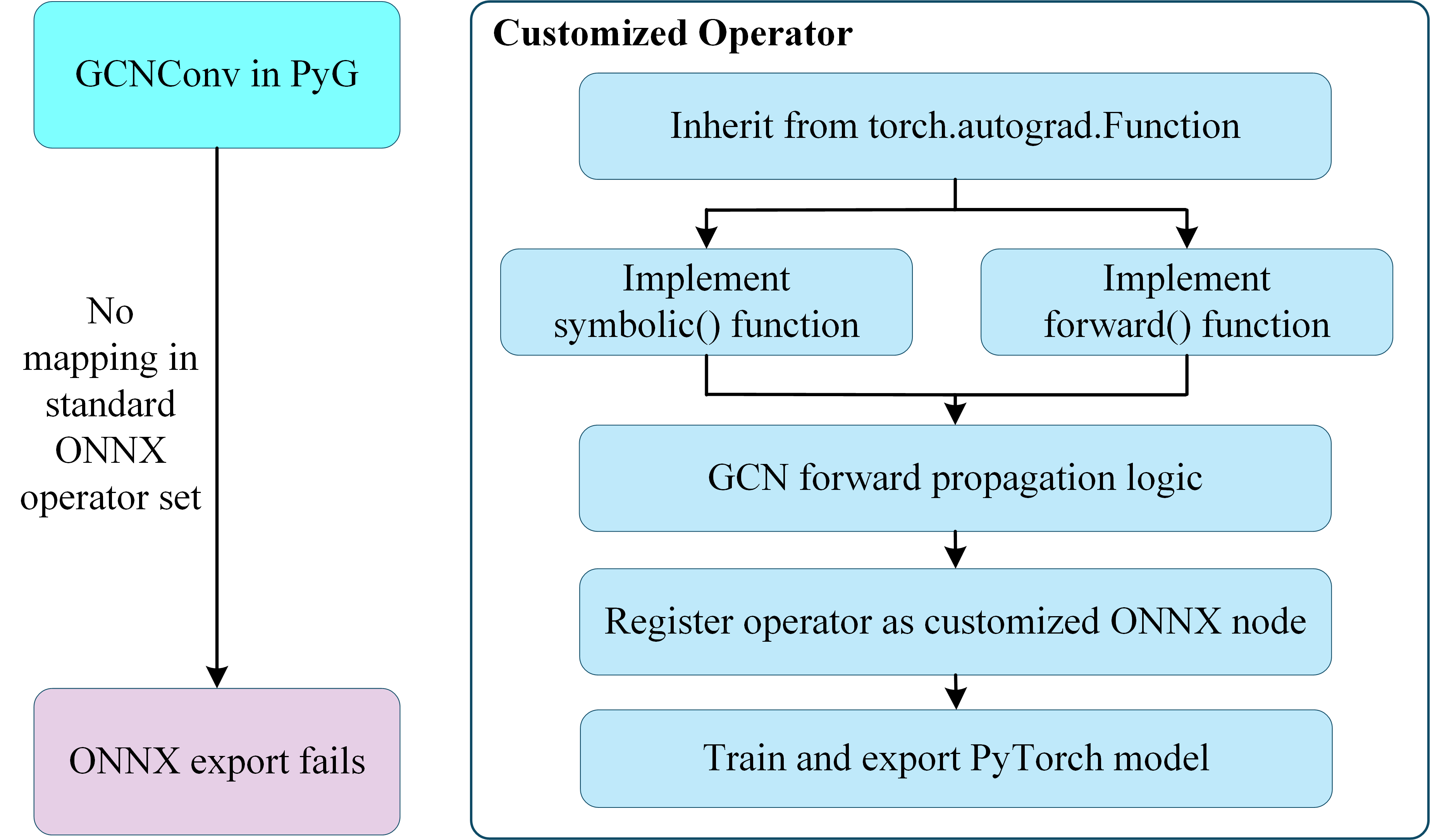}
    \caption{Flowchart regarding how to register the customized operator in ONNX}
    \label{fig:custom_operator_py}
\end{figure}

\begin{figure*}[t]
    \centering
    \resizebox{\textwidth}{!}{%
        \rotatebox{90}{\includegraphics{gcn_onnx_computation_graph.png}}
    }
    \caption{The ONNX's computation graph for the exported GCN model, grouped into functional blocks: Data Input, 2-layer GCN (with MyGcnOp), Flattening/Reshaping, and the MLP prediction head.}
    \label{fig:gcn_onnx}
\end{figure*}

\subsection{Training and Deployment of GraphSAGE}

\begin{figure}
    \centering
    \includegraphics[width=1\linewidth]{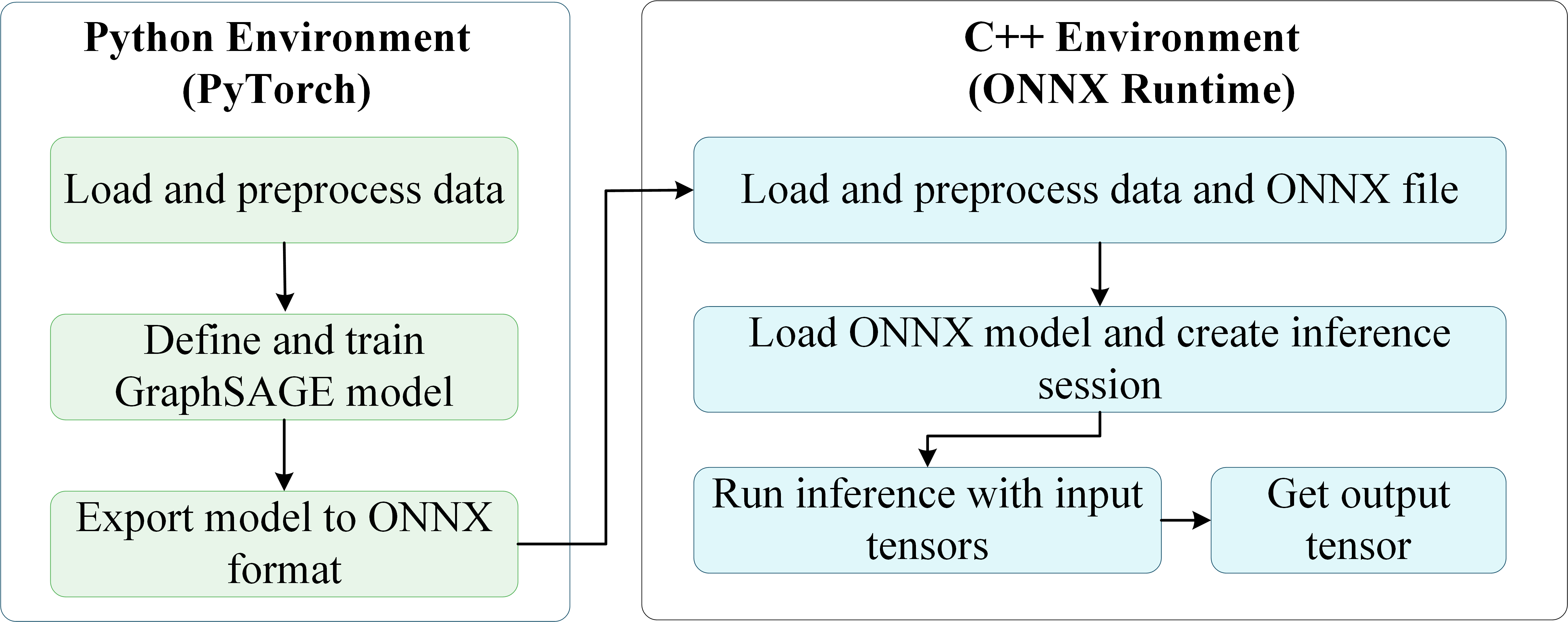}
    \caption{The process for deploying the GraphSAGE model}
    \label{fig:graphsage_deployment}
\end{figure}

Unlike GCN, the GraphSAGE model does not require the development of customized operators. However, if the ONNX exporter is used directly, the resulting ONNX model shows a significant discrepancy in calculation results compared to the original PyTorch model. This is mainly because the ONNX exporter struggles to accurately decompose operators for the complex graph batch-processing operation in GraphSAGE.

To solve this issue, we \textit{rewrote} the batch processing operation in the model's forward function into a serial processing of individual samples, as follows:

\begin{equation}
\begin{split}
    h_{v, i}^k &= \sigma\big(W^k \cdot \text{CONCAT}(h_{v, i}^{k-1}, h_{\mathcal{N}(v), i}^k)\big) \\
    &\quad \text{for } i \in \{1, \dots, B\}
\end{split}
\end{equation}
where:
\begin{itemize}
    \item $h_{v,i}^k$ is the feature of node $v$ for the $i$-th sample in the batch at the $k$-th layer.
    \item $B$ is the batch size.
\end{itemize}

Through this approach, the exported ONNX model maintains a high degree of consistency with PyTorch. After this crucial modification, we only need to follow the standard deployment process (shown in Fig. \ref{fig:graphsage_deployment}) to complete the deployment of the GraphSAGE model on the meter.

\section{Case Study} \label{sec4}

The performance of the proposed method is validated using a field dataset acquired from a real village microgrid. This dataset comprises historical power generation records from three residential PV systems, collected at 15-minute resolution (96 samples daily) from January to May 2024. The goal is to perform a 24-hour-ahead power output forecasting for each PV system. To quantify prediction accuracy, the Mean Absolute Percentage Error (MAPE) is adopted as the primary evaluation metric, as shown in Eq. (\ref{eqn2}), where:
\begin{itemize}
    \item $y_i$ and $\hat{y}_i$ denote the ground truth and 
    the predicted value, respectively;
    \item $N$ represents the total number of test samples;
    \item $Cap$ means the capacity of the PV system.
\end{itemize}

\begin{equation}
\label{eqn2}
\text { MAPE }=\left(1-\sqrt{\frac{1}{N} \sum_{i=1}^N\left(\frac{y_i-\widehat{y_i}}{Cap}\right)^2}\right) \times 100 \%
\end{equation}

\subsection{Results of GCN Model}

We evaluated the model's inference performance with our customized GCN operators on both the PC and the smart meter, as shown in Table \ref{table1}. In this experiment, the data from the three PV systems are consolidated into a feature matrix as the model input, ensuring consistent test sample length and inference time.

The experimental results clearly indicate that our model exhibits high consistency in prediction accuracy on the PC and the meter. The MAPE calculated on the smart meter and on the PC is identical, indicating that our model maintains its predictive accuracy even when deployed on a resource-limited platform like the smart meter.

Note that the models are trained offline on the PC only once. Then, the trained models, with fixed, deterministic weights, are exported to the ONNX format and deployed to the smart meter for inference. Thus, given the same input, the trained model yields identical outputs, resulting in perfect consistency in MAPE across the two hardware platforms.

Furthermore, the results show that the model's runtime on the smart meter is about twice that on the PC, meaning its inference speed on the smart meter is half that on the PC. Although the speed is lower, it can still perform efficient inference for about 2000 data samples per second on the smart meter. This performance is entirely sufficient to meet the requirements of most PV power forecasting tasks.

\begingroup
\renewcommand{\arraystretch}{0.6}
\setlength{\aboverulesep}{1pt}
\setlength{\belowrulesep}{1pt}
\newcommand{\headercell}[1]{{\fontsize{5pt}{6.5pt}\selectfont\makecell{#1}}}

\newcommand{\customtiny}{\fontsize{5pt}{0pt}\selectfont}

\begin{table}[htbp]
\centering
\caption{Results Comparison for the GCN Model}
\label{table1}
\resizebox{\columnwidth}{!}{%
\begin{tabular}{c c| c c |c c}
\toprule
\headercell{\textbf{PV} \\ \textbf{ID}} & 
\headercell{\textbf{Data} \\ \textbf{Length}} & \headercell{\textbf{On-PC} \\ \textbf{Time (sec)}} & \headercell{\textbf{MAPE}} & \headercell{\textbf{On-Meter} \\ \textbf{Time (sec)}} & \headercell{\textbf{MAPE}}\\
\midrule
{\customtiny1} & {\customtiny
2209} & {\customtiny0.627} & {\customtiny9.01\%} & {\customtiny1.199} & {\customtiny9.01\%}\\
{\customtiny2} & {\customtiny2209} & {\customtiny0.627} & {\customtiny8.42\%} & {\customtiny1.199} & {\customtiny8.42\%}\\
{\customtiny3} & {\customtiny2209} & {\customtiny0.627} & {\customtiny8.99\%} & {\customtiny1.199} & {\customtiny8.99\%}\\

\bottomrule
\end{tabular}
}
\end{table}
\endgroup

\subsection{Results of GraphSAGE Model}

\begin{figure}
    \centering
    \includegraphics[width=1\linewidth]{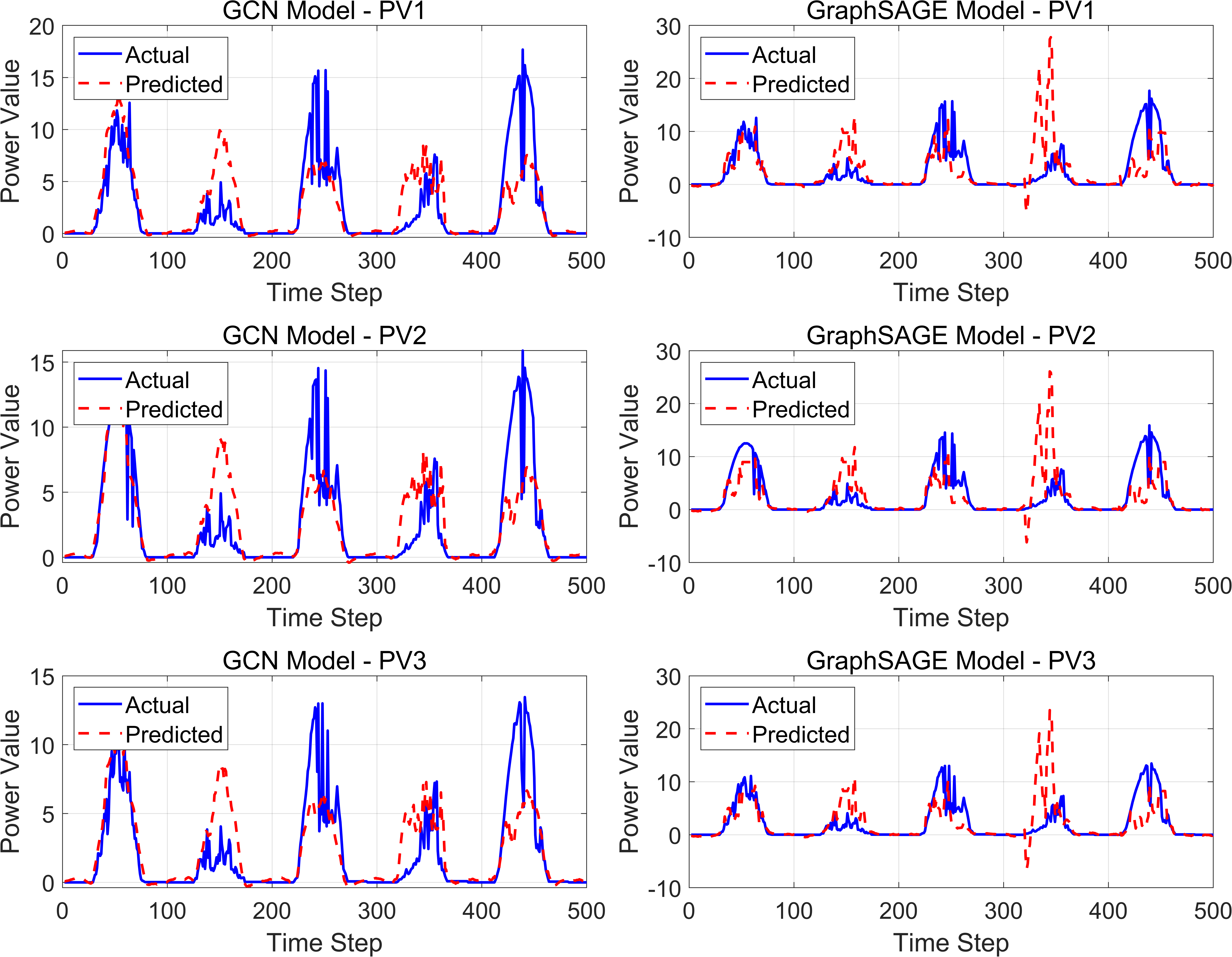}
    \caption{Comparison of on-meter forecasting results: GCN vs. GraphSAGE}
    \label{fig:resDisplay}
\end{figure}

We also evaluate GraphSAGE's performance across different platforms, as shown in Table \ref{table2}. Fig \ref{fig:resDisplay} compares the performance of the kinds of models on the first 500 test cases for the three photovoltaic users. The test samples are the same as those used for the GCN model. As the table shows, the prediction accuracy of our GraphSAGE model is consistent across different platforms, which once again verifies the model's stability for cross-platform deployment.

Comparing the two models, the GraphSAGE model's inference speed is significantly faster than the GCN model on the PC. However, on the meter, the GraphSAGE model's inference speed is slower than that of the GCN model. In terms of prediction accuracy, the GCN model's overall MAPE is lower than that of the GraphSAGE model, indicating higher accuracy for the GCN model in this test. 

It is likely that our input graph is small, and because GraphSAGE relies more on local sampling and aggregation mechanisms (which lead to information loss in graphs with fewer nodes), it has difficulty in learning optimal feature representations. In contrast, on such a small-scale graph, GCN can fully leverage all available latent information from neighbors, enabling it to better capture node relationships. 

To evaluate the feasibility of model deployment on the meter, we also measure resource usage during model execution. The GCN model requires 121.3MB peak memory and 235\% CPU, indicating efficient multi-thread execution. In contrast, GraphSAGE consumes 215.4MB memory and 99\% CPU, suggesting single-core execution. Overall, both models use only a small portion of memory, confirming they are sufficiently lightweight for continuous on-meter inference.

In summary, on the meter, the GCN model is superior in accuracy and more efficient in inference speed.

\begingroup
\renewcommand{\arraystretch}{0.6}
\setlength{\aboverulesep}{1pt}
\setlength{\belowrulesep}{1pt}
\newcommand{\headercell}[1]{{\fontsize{5pt}{6.5pt}\selectfont\makecell{#1}}}

\newcommand{\customtiny}{\fontsize{5pt}{0pt}\selectfont}

\begin{table}[htbp]
\centering
\caption{Results Comparison for the GraphSAGE Model}
\label{table2}
\resizebox{\columnwidth}{!}{%
\begin{tabular}{c c| c c |c c}
\toprule
\headercell{\textbf{PV} \\ \textbf{ID}} & 
\headercell{\textbf{Data} \\ \textbf{Length}} & \headercell{\textbf{On-PC} \\ \textbf{Time (sec)}} & \headercell{\textbf{MAPE}} & \headercell{\textbf{On-Meter} \\ \textbf{Time (sec)}} & \headercell{\textbf{MAPE}}\\
\midrule
{\customtiny1} & {\customtiny
2209} & {\customtiny0.118} & {\customtiny10.29\%} & {\customtiny2.970} & {\customtiny10.29\%}\\
{\customtiny2} & {\customtiny2209} & {\customtiny0.118} & {\customtiny9.29\%} & {\customtiny2.970} & {\customtiny9.29\%}\\
{\customtiny3} & {\customtiny2209} & {\customtiny0.118} & {\customtiny10.33\%} & {\customtiny2.970} & {\customtiny10.33\%}\\

\bottomrule
\end{tabular}
}
\end{table}
\endgroup

\section{Conclusions}
In this paper, we have proposed the development process of a customized GCN operator for edge devices and successfully deployed both GCN and GraphSAGE models on a smart meter for PV forecasting. The test results show that the model's MAPE remained consistent between the PC and the smart meter, which demonstrates the cross-platform stability of our models. In the future, we will explore how to apply these technologies to additional types of meters and develop more advanced graph machine learning operators.

\section*{Acknowledgment}

The author would like to thank the support from the ``Ministry of Education-Huawei Industry-Education Integration Smart Base 2.0" project (SYSU-76150-20240820-0002).

\end{document}